\documentclass{bmvc2k}

\usepackage{microtype}
\usepackage{wrapfig}
\usepackage{graphicx}
\usepackage[table]{xcolor}
\usepackage{booktabs}
\usepackage{amsfonts}
\usepackage{multirow}
\usepackage{cleveref}
\usepackage{float}
\usepackage{amssymb}
\usepackage{threeparttable}
\usepackage{pifont}
\newcommand{\cmark}{\ding{51}}
\newcommand{\xmark}{\ding{55}}
\title{HSTGFormer: Hyper Spatial-Temporal Graph Transformer for 3D Human Pose Estimation}

\addauthor{Ruochen Li}{ruochen.li@durham.ac.uk}{1}
\addauthor{Shuang Chen}{shuang.chen@durham.ac.uk}{1}
\addauthor{Wenke E}{wenke.e@durham.ac.uk}{1}
\addauthor{Farshad Arvin}{farshad.arvin@durham.ac.uk}{1}
\addauthor{Amir Atapour-Abarghouei}{amir.atapour-abarghouei@durham.ac.uk}{1}

\usepackage{amsfonts}
\addinstitution{
 Department of Computer Science\\Durham University\\Durham, UK
}

\runninghead{LI, CHEN, E, ARVIN, ATAPOUR-ABARGHOUEI}{HSTGFormer}

\begin{document}

\maketitle

\begin{abstract}

Transformer-based methods have achieved strong performance in monocular 3D human pose estimation, but most existing approaches organise spatial and temporal reasoning as separate stages, which may weaken unified spatial-temporal interdependencies inherent in human motion and compress frame-level structural information before temporal modelling. In this paper, we propose HSTGFormer, a graph-enhanced Transformer framework that reformulates spatial-temporal reasoning as localised coupled graph aggregation over joint-time nodes. Specifically, HSTGFormer introduces a Hyper Spatial-Temporal Graph (HSTG), which decomposes global spatial-temporal reasoning into local spatial-temporal receptive fields around individual joint-time nodes by extending per-frame skeleton graphs into temporal neighbourhoods, thereby enabling structure-aware coupled reasoning while preserving local structural motion information. It further incorporates an Adaptive Dual-Scale Temporal Graph (ADSTG) to capture joint-specific temporal dependencies over complementary short- and long-range windows. A lightweight node-wise fusion module further adaptively integrates the two graph representations for each joint-time node. Experiments on Human3.6M and MPI-INF-3DHP show that HSTGFormer achieves strong accuracy with high computational efficiency.
\end{abstract}

%-------------------------------------------------------------------------
\section{Introduction}
\label{sec:intro}

3D human pose estimation (HPE) aims to recover accurate human joint coordinates from input monocular 2D images, which is a crucial task in computer vision with broad research significance and related applications such as motion prediction \cite{yang20223d,cui2023test}, human-computer interaction \cite{rossol2015multisensor,huo20233d,qiao2025geometricGNN,qiao2024category}, sports analysis \cite{suzuki2025sportanalysis,ingwersen2023sportspose}, and virtual reality \cite{hagbi2010shape}. Unlike single-frame pose estimation \cite{li2022mhformer,zhao2023poseformerv2,yu2023glagcn}, video-based methods exploit both the anatomical structure of the human body and the temporal dynamics of motion. Skeleton provides strong kinematic constraints among connected joints, while consecutive frames contain motion cues that help resolve depth ambiguity, occlusion, and noisy detections. Therefore, effective 3D HPE requires comprehensive modelling of the spatial-temporal correlations of human motions.

\begin{figure}[t!]
\centering
  \includegraphics[width=0.93\linewidth]{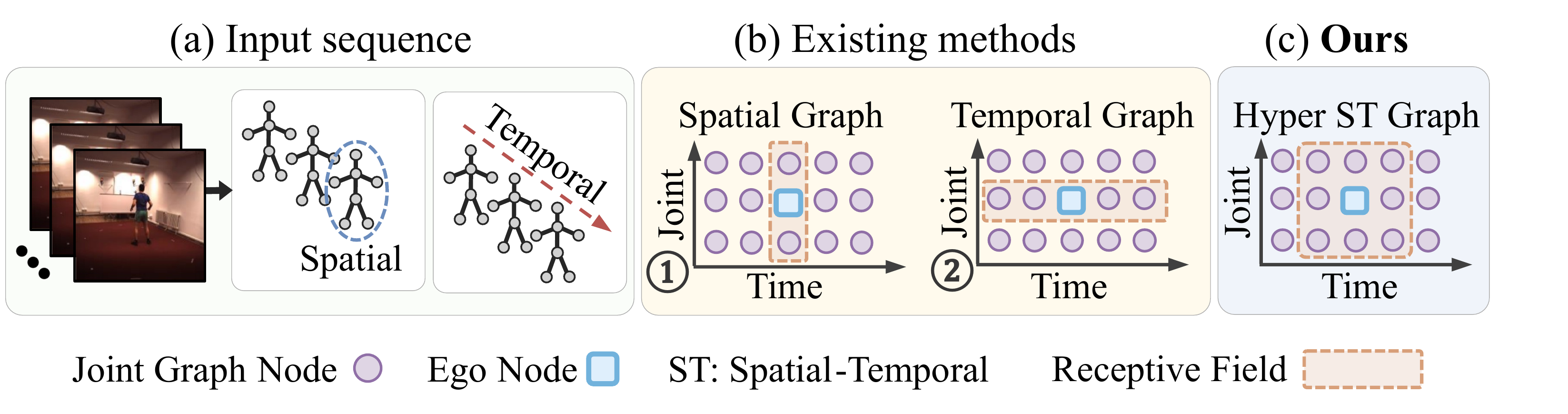}
\caption{
Illustration of the proposed \textit{Hyper Spatial-Temporal Graph} (HSTG).
\textbf{(a)} 2D joints are represented as graph nodes over time.
\textbf{(b)} Existing methods decouple spatial and temporal reasoning through intra-frame skeletal aggregation followed by same-joint temporal modelling.
\textbf{(c)} HSTG defines a localised coupled spatial-temporal receptive field around each ego node, efficiently aggregating from anatomically related joints within nearby frames.
}
\label{fig:teaser}
\end{figure}

Early methods for 3D human pose estimation typically operated on individual frames and adopted CNN-based architectures to either directly regress 3D joint coordinates from images \cite{li20143dv1} or to lift 2D keypoint detections to 3D poses using fully connected networks \cite{martinez2017singlesimple}. These methods generally lack effective and explicit spatial-temporal dependency modelling, relying primarily on per-frame predictions with limited structural priors. Inspired by the representation learning capability of Transformers \cite{vaswani2017attention,devlin2019bert} in natural language processing, many video-based 3D HPE methods have adopted attention mechanisms to model spatial-temporal dependencies \cite{zhu2023motionbert,zheng2021poseformer,tang2023stcformer,peng2024ktpformer,mehraban2024motionagformer,liu2025tcpformer,zhang2022mixste}. To model spatial-temporal correlations, these methods employ spatial attention to capture interactions among joints within each frame, while using temporal attention to model motion dependencies across consecutive frames. Despite their strong performance, Transformer-based methods often lack explicit skeletal priors and require stacked attention layers to capture complex spatial-temporal correlations, resulting in high computational cost for long video sequences. To address these limitations, recent works \cite{peng2024ktpformer,mehraban2024motionagformer,li2026MASC} incorporate graphs (Fig.~\ref{fig:teaser}(a)) into Transformer-based frameworks to introduce explicit skeletal priors and improve spatial-temporal representation learning. By modelling human joints as graph nodes, graph-based methods provide stronger structural inductive bias and enable efficient representation learning.

Despite recent progress, existing methods still face several limitations in spatial-temporal modelling. First, most existing methods organise spatial and temporal reasoning as two separate propagation stages, where spatial features are first extracted within each frame and subsequently passed to temporal modules for motion modelling (Fig.~\ref{fig:teaser}(b)). This separate process compresses frame-level structural information before temporal reasoning, limiting the ability to capture continuous spatial-temporal interdependencies \cite{yi2023fouriergnn,li2025unified,marisca2025oversquashing,li2026vite,li2025bpsgcn}. For example, during walking, the knee and foot not only exhibit strong intra-frame anatomical coupling, but also maintain correlated local motion patterns across adjacent timesteps. However, sequential spatial-then-temporal propagation may weaken such localised spatial-temporal dependencies before coupled reasoning is established. Second, temporal modelling is often conducted over either long-range sequences or predefined receptive fields. Although recent works explore multi-scale temporal modelling~\cite{liu2025tcpformer}, they still rely on fixed temporal scales, which may be insufficient for modelling joint dynamics across different motion patterns (e.g., fast-moving extremities vs.\ relatively stable torso joints).

In this paper, we propose \textbf{HSTGFormer}, a graph-based Transformer framework for efficient 3D human pose estimation. It introduces a graph-enhanced spatial-temporal reasoning paradigm based on two complementary graph structures: \textbf{Hyper Spatial-Temporal Graph} (HSTG) and \textbf{Adaptive Dual-Scale Temporal Graph} (ADSTG). As shown in Fig.~\ref{fig:teaser}(c), HSTG moves beyond decoupled per-frame spatial and per-joint temporal receptive fields by extending skeleton graphs into localised temporal neighbourhoods, yielding a local coupled spatial-temporal receptive field for each joint-time node. This formulation preserves localized spatial-temporal interdependencies within each joint-time neighbourhood, alleviating early information compression \cite{marisca2025oversquashing} of frame-level skeletal features before temporal modelling. Through efficient graph-based aggregation, HSTG preserves anatomical priors and captures local spatial-temporal context without constructing a dense spatial-temporal graph. For temporal modelling, ADSTG constructs dynamic graphs over short- and long-range temporal windows. This allows each joint to adaptively aggregate motion dependencies from different temporal contexts, enabling flexible temporal reasoning for diverse joint motion patterns. Finally, a context-aware fusion mechanism assigns node-wise weights to HSTG and ADSTG, allowing each node to integrate complementary spatial-temporal contexts.

Experiments on Human3.6M~\cite{h36m} and MPI-INF-3DHP~\cite{3dhp} show that HSTGFormer achieves strong accuracy with high computational efficiency. Our contributions are: 
\begin{itemize}
    \item We propose \textbf{HSTGFormer}, a graph-based Transformer framework for efficient 3D human pose estimation, which reformulates spatial-temporal modelling from a graph perspective to better capture motion continuity and structural dependencies.
    \item We introduce a \textit{Hyper Spatial-Temporal Graph (HSTG)} structure (Sec.~\ref{sec:hstg}) that extends per-frame skeleton graphs to localised temporal neighbourhoods, enabling structure-aware modelling of local motion continuity while preserving human skeletal priors.
    \item We design an \textit{Adaptive Dual-Scale Temporal Graph (ADSTG)} module (Sec.~\ref{sec:adstg}) to construct and fuse dynamic temporal graphs over complementary temporal ranges, allowing joint-specific and adaptive temporal dependency modelling.
\end{itemize}

\section{Related Work}
\label{sec:related}

\subsection{Transformer for 3D Human Pose Estimation}

With the success of Transformers \cite{vaswani2017attention} in natural language processing \cite{devlin2019bert} and computer vision \cite{dosovitskiy2020vit,li2026art}, recent works have increasingly adopted Transformer-based architectures for 3D human pose estimation. PoseFormer \cite{zheng2021poseformer} introduces one of the first pure Transformer frameworks for 3D pose lifting, where spatial and temporal Transformers are stacked to jointly model spatial relationships within each frame and long-range dependencies across frames. MixSTE \cite{zhang2022mixste} proposes a mixed spatial-temporal encoder with an alternating seq2seq design that more effectively captures temporal dynamics of different body joints over long sequences. MotionBERT \cite{zhu2023motionbert} employs large-scale pre-training to learn general motion representations from massive pose data and adapts them to downstream 3D pose estimation tasks.
STCFormer \cite{tang2023stcformer} introduces a two-pathway design for spatial-temporal modelling, enabling more expressive modeling of complex motion patterns in monocular videos. TCPFormer \cite{liu2025tcpformer} further refines temporal context aggregation by tailoring Transformer blocks to better capture multi-scale temporal cues and mitigate motion ambiguity, leading to consistent performance gains on standard 3D pose benchmarks. However, most Transformer-based methods do not explicitly exploit human skeletal priors, often requiring deeper architectures to model complex spatial-temporal dependencies, which increases computational cost.

\subsection{Graph-transformer for 3D Human Pose Estimation}
Due to the strong capability for modelling topological structures, graph-based methods have been widely adopted for spatial-temporal modelling in 3D human pose estimation. 
PoseGTAC~\cite{zhu2021posegtac} combines graph atrous convolution for multi-scale neighbour aggregation with graph Transformer layers to capture long-range spatial-temporal dependencies. 
MotionAGFormer \cite{mehraban2024motionagformer} integrates attention and GCN \cite{kipf2016GCN,ruochen2022multiclassSGCN} branches to capture global joint relationships and local skeletal dependencies. 
DiffPose \cite{gong2023diffpose} further introduces a diffusion-based framework with GCN  to better encode spatial dependencies on the human skeleton and generate more reliable multi-hypothesis 3D poses under occlusion.
GLA-GCN \cite{yu2023glagcn} explores adaptive graph convolution to capture global pose representations. By adopting a strided temporal design, it reduces the effective temporal modelling range and achieves competitive performance compared with Transformer-based methods, while requiring lower memory consumption.
KTPFormer \cite{peng2024ktpformer} introduces kinematic and trajectory priors into spatial and temporal attention to better encode anatomical and motion constraints. However, most existing graph-transformer methods still perform spatial modelling within individual frames, limiting their ability to explicitly capture continuous spatial-temporal interdependencies during human motion.

\section{Method}
\label{sec:method}

\subsection{Problem Formulation and Overview}
Given an input 2D pose sequence $\mathbf{X}=\{\mathbf{x}_t\}_{t=1}^{T} \in \mathbb{R}^{T \times J \times C_{\text{in}}}$, where $T$ is the number of frames and $J$ is the number of body joints, each joint is represented by its 2D coordinates and confidence score. The objective is to recover the corresponding 3D pose sequence $\mathbf{Y}=\{\mathbf{y}_t\}_{t=1}^{T} \in \mathbb{R}^{T \times J \times C_{\text{out}}}$, where $C_{\text{in}}$ and $C_{\text{out}}$ denote the input and output joint dimensions, respectively. Following recent Transformer-based 3D pose estimation methods~\cite{liu2025tcpformer,mehraban2024motionagformer}, we first map the input sequence into a high-dimensional latent space through a linear embedding layer, producing $\mathbf{X}_{h} \in \mathbb{R}^{T \times J \times D}$, where $D$ is the hidden feature dimension. To retain joint identity information, a learnable joint-level positional encoding $\mathbf{P}_{\text{pos}}$ is added to the embedded features. The resulting representation is then processed by a spatial-temporal Transformer encoder $ \mathbf{X}_{st} = \Phi\left(\mathbf{X}_{h} + \mathbf{P}_{\text{pos}}\right)$,
where $\Phi(\cdot)$ denotes the backbone encoder used to capture global spatial-temporal dependencies before graph-based reasoning.

\begin{figure}[t]
\centering
  \includegraphics[width=0.95\linewidth]{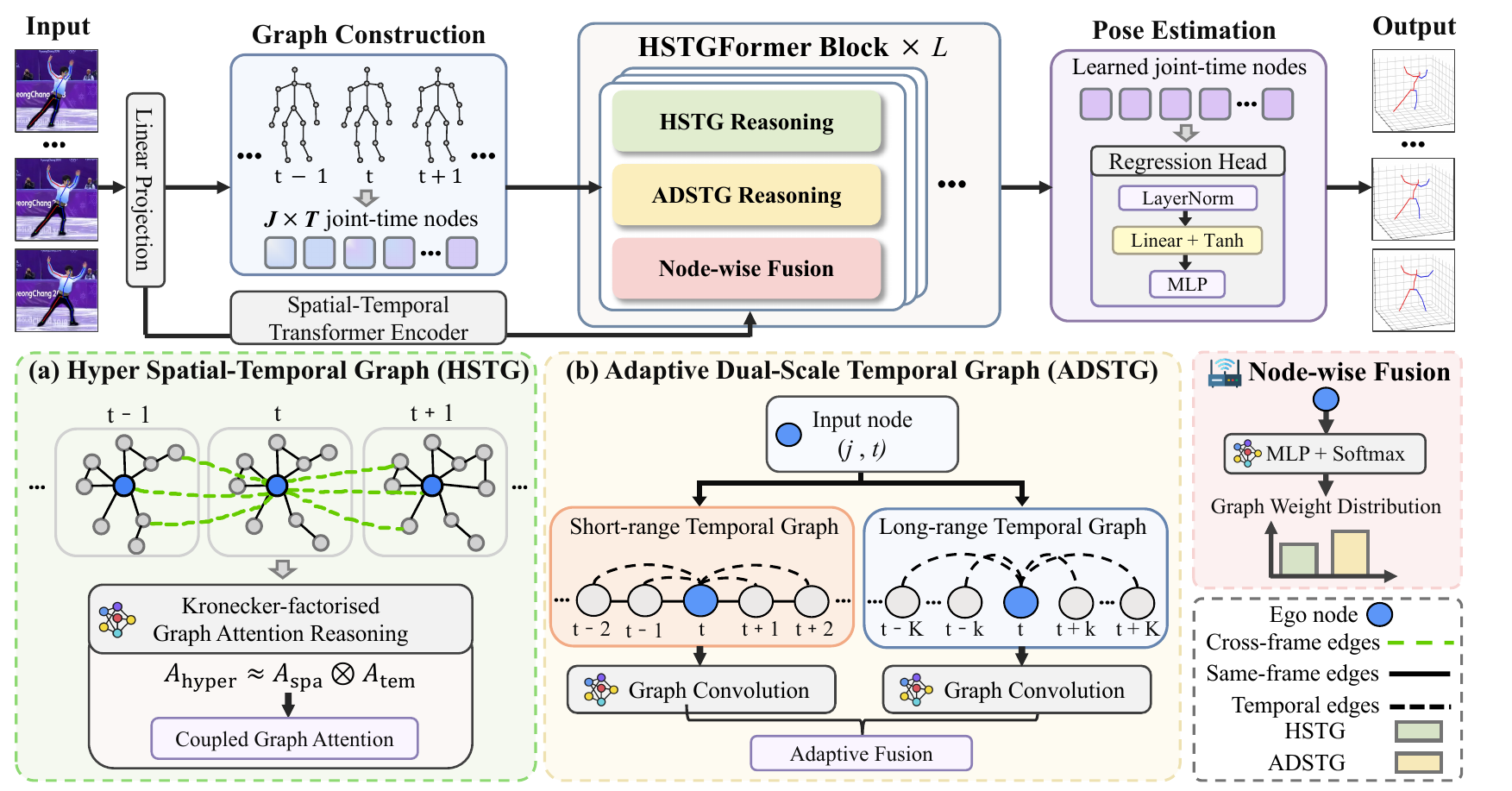}
  \caption{Framework overview. HSTGFormer consists of two graph-based modules: (a) the Hyper Spatial-Temporal Graph (HSTG), which constructs localised coupled spatial-temporal receptive fields across nearby frames, and (b) the Adaptive Dual-Scale Temporal Graph (ADSTG), which models adaptive short- and long-range temporal dependencies. A node-wise fusion module then adaptively integrates graph contexts for final 3D pose estimation.}
  \label{fig:overview}
\end{figure}

As illustrated in Fig.~\ref{fig:overview}, building on the embedded pose representation $\mathbf{X}_{h}$, HSTGFormer performs graph-enhanced spatial-temporal reasoning through two complementary branches. The first branch, \textit{Hyper Spatial-Temporal Graph} (HSTG), extends the per-frame skeleton graph to localised temporal neighbourhoods, forming a coupled spatial-temporal receptive field for each joint-time node. Through mask-constrained and factorised graph attention, HSTG captures local structural motion context while preserving anatomical connectivity priors. The second branch, \textit{Adaptive Dual-Scale Temporal Graph} (ADSTG), constructs dynamic graphs over short- and long-range temporal windows, allowing each joint to adaptively aggregate motion cues from complementary temporal ranges. A context-aware node-wise fusion module then predicts adaptive weights for the two branches, enabling each joint-time node to integrate local spatial-temporal context and temporal motion dynamics. The fused graph representation is finally combined with the backbone features and fed into the regression head to estimate the 3D pose sequence.

\subsection{Hyper Spatial-Temporal Graph (HSTG) Reasoning}
\label{sec:hstg}

Human motion is governed by anatomical constraints and local temporal continuity, but conventional spatial-then-temporal methods model them separately: first within each frame, then along each joint trajectory. Such designs may overlook localised spatial-temporal interdependencies, where neighbouring joints exhibit correlated motion over adjacent timesteps. 
To address this, we introduce the \textit{Hyper Spatial-Temporal Graph} (HSTG), which extends the skeleton graph from individual frames to local temporal neighbourhoods and performs structure-aware graph reasoning over joint-time nodes.

\paragraph{Graph Construction.}
Given the embedded pose representation $\mathbf{X}_{h}\in\mathbb{R}^{T\times J\times D}$, we construct a Hyper Spatial-Temporal Graph $\mathcal{G}_{\text{hyper}}=(\mathcal{V}, \mathcal{E}_{\text{hyper}})$, where each node $v_{t,j}\in\mathcal{V}$ represents joint $j$ at timestep $t$, and $|\mathcal{V}|=TJ$. 
From the perspective of each ego node $v_{t,j}$, HSTG defines a local spatial-temporal neighbourhood by jointly considering anatomical connectivity and temporal proximity. 
Specifically, let $\mathbf{A}_{\text{spa}}\in\{0,1\}^{J\times J}$ denote the skeleton adjacency matrix with self-loops, and let $\mathbf{A}_{\text{tem}}\in\{0,1\}^{T\times T}$ denote a temporal band adjacency:
\begin{equation}
[\mathbf{A}_{\text{tem}}]_{t,t'}=
\begin{cases}
1, & |t-t'|\leq w,\\
0, & \text{otherwise},
\end{cases}
\end{equation}
where $w$ is the temporal window radius.
The resulting spatial-temporal support can be represented in a Kronecker-factorised form:
\begin{equation}
\mathbf{A}_{\text{hyper}} \approx \mathbf{A}_{\text{tem}}\otimes \mathbf{A}_{\text{spa}},
\end{equation}
where $\otimes$ denotes the Kronecker product. 
Here, $\mathbf{A}_{\text{hyper}}$ denotes the induced local spatial-temporal support rather than an explicitly materialised dense adjacency matrix. 
This factorised formulation allows HSTG to perform graph attention through spatial and temporal graph operations while preserving a localised coupled spatial-temporal receptive field.

\paragraph{Factorised Graph Attention.}
The support $\mathbf{A}_{\text{hyper}}$ specifies the valid local spatial-temporal neighbourhood of each joint-time node. 
However, explicitly performing attention over $\mathbf{A}_{\text{hyper}}$ would require constructing a $(TJ)\times(TJ)$ attention map, which is computationally expensive for long pose sequences. 
Instead, we exploit the Kronecker structure of $\mathbf{A}_{\text{hyper}}$ and implement HSTG reasoning through two factorised masked graph attention \cite{velivckovic2017gat} operations: skeleton-constrained spatial attention followed by local temporal attention.

Given a set of node features $\mathbf{F}\in\mathbb{R}^{N\times D}$ and a binary adjacency matrix $\mathbf{A}\in\{0,1\}^{N\times N}$, we define a generic masked graph attention operator~\cite{velivckovic2017gat,vaswani2017attention} as
\begin{equation}
\begin{gathered}
\mathrm{GraphAttn}(\mathbf{F}, \mathbf{A})_{i}
=
\sum_{j\in\mathcal{N}(i)}
\alpha_{ij}\mathbf{F}_{j}\mathbf{W}_{v}, \\
\alpha_{ij}
=
\frac{
\exp\left(
(\mathbf{F}_{i}\mathbf{W}_{q})
(\mathbf{F}_{j}\mathbf{W}_{k})^{\top}/\sqrt{D}
\right)
}{
\sum_{k\in\mathcal{N}(i)}
\exp\left(
(\mathbf{F}_{i}\mathbf{W}_{q})
(\mathbf{F}_{k}\mathbf{W}_{k})^{\top}/\sqrt{D}
\right)
},
\end{gathered}
\end{equation}
where $\mathcal{N}(i)=\{j\mid \mathbf{A}_{ij}=1\}$ denotes the masked neighbourhood of node $i$, and $\mathbf{W}_{q}$, $\mathbf{W}_{k}$, and $\mathbf{W}_{v}$ are learnable projection matrices.

Using this operator, we first normalise \cite{ba2016layernorm} the embedded feature as $
\bar{\mathbf{X}}_{h}=\mathrm{LN}(\mathbf{X}_{h}).
$
For each timestep $t$, we apply skeleton-constrained graph attention over the joint dimension:
\begin{equation}
\mathbf{Z}_{t}
=
\mathrm{GraphAttn}_{\mathrm{spa}}
\left(
\bar{\mathbf{X}}^{t}_{h}, 
\mathbf{A}_{\mathrm{spa}}
\right),
\end{equation}
where $\bar{\mathbf{X}}^{t}_{h}\in\mathbb{R}^{J\times D}$ denotes the joint features at timestep $t$. 
Then, for each joint $j$, temporal graph attention is applied over its local temporal neighbourhood:
\begin{equation}
\mathbf{H}^{j}
=
\mathrm{GraphAttn}_{\mathrm{tem}}
\left(
\mathbf{Z}^{j}, 
\mathbf{A}_{\mathrm{tem}}
\right),
\end{equation}
where $\mathbf{Z}^{j}\in\mathbb{R}^{T\times D}$ denotes the temporal sequence of the spatially enriched features for joint $j$. Finally, we apply residual refinement to obtain the HSTG output:
\begin{equation}
\mathbf{X}_{\text{hyper}}
=
\mathbf{X}_{h}
+
\sigma\left(
\mathrm{LN}
\left(
\mathbf{H}+\bar{\mathbf{X}}_{h}\mathbf{W}_{u}
\right)
\right),
\end{equation}
where $\sigma(\cdot)$ denotes the ReLU activation.

HSTG defines local joint-time receptive fields instead of sequentially separating spatial and temporal modelling. Its factorised attention preserves the localised Cartesian support between skeleton and temporal neighbours, enabling efficient spatial-temporal graph reasoning without constructing a dense attention graph.

\subsection{Adaptive Dual-Scale Temporal Graph (ADSTG) Reasoning}
\label{sec:adstg}

Although HSTG captures localised coupled spatial-temporal context, human motion also involves temporal dependencies at different ranges. For example, fast-moving joints require short-range cues for fine-grained motion changes, while more stable joints may benefit from longer-range context. In this paper, we propose an \textit{Adaptive Dual-Scale Temporal Graph} (ADSTG), which constructs dynamic temporal graphs over complementary short- and long-range windows and fuses them in a joint-aware manner.

\paragraph{Dynamic Temporal Graph Construction.}
Given the embedded pose representation $\mathbf{X}_{h}\in\mathbb{R}^{T\times J\times D}$, ADSTG constructs dynamic temporal graphs along the temporal dimension. 
Unlike HSTG, which defines a localised coupled spatial-temporal receptive field, ADSTG focuses on selecting informative temporal neighbours for each joint-time node $(t,j)$ from complementary temporal ranges. 
Specifically, we define two centred temporal windows:
\begin{equation}
\mathcal{N}_{m}(t)=\{t'\mid |t-t'|\leq r_m\},
\qquad m\in\{\mathrm{short},\mathrm{long}\},
\end{equation}
where $r_m$ denotes the temporal radius of branch $m$. 
Within each window, ADSTG computes dot-product similarities between node $(t,j)$ and its temporal candidates $(t',j)$, and retains the top-$k_m$ most similar candidates to form a sparse dynamic temporal adjacency:
\begin{equation}
[\mathbf{A}_{m}]_{t,t'}=
\begin{cases}
1, & t'\in \mathrm{TopK}_{k_m}\!\left(\mathcal{N}_{m}(t)\right),\\
0, & \text{otherwise},
\end{cases}
\end{equation}
where $\mathbf{A}_{\mathrm{short}}$ and $\mathbf{A}_{\mathrm{long}}$ denote the short- and long-range dynamic temporal adjacencies, respectively. 
This construction allows each joint-time node to adaptively select relevant temporal neighbours within local and extended temporal contexts.

\paragraph{Dual-Scale Temporal Propagation.}
Given a dynamic temporal adjacency $\mathbf{A}$, we define a normalised temporal graph GCN~\cite{kipf2016GCN} as
\begin{equation}
\begin{gathered}
\mathrm{TempGCN}(\mathbf{X}_{h}, \mathbf{A})
=
\tilde{\mathbf{A}}\mathbf{X}_{h}\mathbf{W}_{v}, \\
\tilde{\mathbf{A}}
=
\mathbf{D}^{-\frac{1}{2}}
(\mathbf{A}+\mathbf{I})
\mathbf{D}^{-\frac{1}{2}},
\end{gathered}
\end{equation}
where self-loops are added before symmetric normalisation, $\mathbf{D}$ is the degree matrix, and $\mathbf{W}_{v}$ is a learnable projection matrix. ADSTG applies the same temporal GCN to the short- and long-range dynamic temporal adjacencies:
\begin{equation}
\mathbf{Z}_{\mathrm{short}}
=
\mathrm{TempGCN}(\mathbf{X}_{h}, \mathbf{A}_{\mathrm{short}}),
\qquad
\mathbf{Z}_{\mathrm{long}}
=
\mathrm{TempGCN}(\mathbf{X}_{h}, \mathbf{A}_{\mathrm{long}}).
\end{equation}
This produces two complementary temporal representations, corresponding to local motion details and extended temporal context.

To combine the two temporal scales, we use a lightweight topology-aware gate based on the skeleton degree. 
Given $\mathbf{A}_{\text{spa}}$, we compute the joint degree vector $\mathbf{d}\in\mathbb{R}^{J}$ and generate joint-wise scale weights $\mathbf{G}$. The output of ADSTG is obtained by weighted fusion of the short- and long-range temporal representations:
\begin{equation}
\mathbf{X}_{\mathrm{ada}}
=
\sum_{m\in\{\mathrm{short},\mathrm{long}\}}
\mathbf{G}_{m}\odot \mathbf{Z}_{m}.
\end{equation}

ADSTG complements HSTG by focusing on adaptive temporal dependency modelling. 
Instead of relying on a single fixed temporal neighbourhood, ADSTG constructs content-adaptive temporal graphs within short- and long-range windows, allowing each joint to select informative temporal neighbours according to its motion pattern. 
The joint-adaptive fusion further balances local and extended temporal contexts, providing flexible temporal reasoning for heterogeneous human motion dynamics.

\subsection{Adaptive Node-wise Fusion}
\label{sec:nodewise_fusion}

After obtaining the HSTG feature $\mathbf{X}_{\mathrm{hyper}}$ and the ADSTG feature $\mathbf{X}_{\mathrm{ada}}$, we use two lightweight branch-specific Multilayer Perceptrons (MLP) for feature enhancement:
\begin{equation}
\hat{\mathbf{X}}_{\mathrm{hyper}}=\mathrm{MLP}_{\mathrm{hyper}}(\mathbf{X}_{\mathrm{hyper}}),
\qquad
\hat{\mathbf{X}}_{\mathrm{ada}}=\mathrm{MLP}_{\mathrm{ada}}(\mathbf{X}_{\mathrm{ada}}).
\end{equation}

To adaptively combine the two graph representations, we introduce a context-aware fusion module that predicts node-wise weights for each joint-time node $(t,j)$. Here, $\mathbf{C}_{\mathrm{spa}}$ is obtained by averaging $\mathbf{X}_{h}$ over joints to provide frame-level context, while $\mathbf{C}_{\mathrm{tem}}$ is obtained by averaging over time to provide joint-level context. After broadcasting them to each joint-time node, we concatenate the token feature with $\mathbf{C}_{\mathrm{spa}}$ and $\mathbf{C}_{\mathrm{tem}}$, and feed the result into a lightweight MLP followed by softmax:

\begin{equation}
\omega_{\mathrm{hyper}},\omega_{\mathrm{ada}}
=
\mathrm{Softmax}
\left(
\mathrm{MLP}
\left[
\mathbf{X}_{h},
\mathbf{C}_{\mathrm{spa}},
\mathbf{C}_{\mathrm{tem}}
\right]
\right).
\end{equation}
The final graph representation is then computed as
\begin{equation}
\mathbf{X}_{\mathrm{fuse}}
=
\omega_{\mathrm{hyper}}\odot \hat{\mathbf{X}}_{\mathrm{hyper}}
+
\omega_{\mathrm{ada}}\odot \hat{\mathbf{X}}_{\mathrm{ada}}.
\end{equation}
This adaptive fusion allows each joint-time node to balance local coupled spatial-temporal reasoning from HSTG and adaptive temporal dependency modelling from ADSTG. The fused representation is then combined with the backbone Transformer encoder features \cite{liu2025tcpformer,mehraban2024motionagformer}, enabling the model to retain global spatial-temporal context while incorporating graph-enhanced motion representations. After $L$ stacked HSTGFormer blocks, an MLP-based regression head maps the learned joint-time features to the final 3D pose sequence.

\subsection{Overall Learning Objectives}

We optimise HSTGFormer end-to-end using a composite training objective following the commonly used protocol in recent 3D HPE methods~\cite{liu2025tcpformer}. The overall loss is defined as:
\begin{equation}
\mathcal{L}
=
\mathcal{L}_{\mathrm{mpjpe}}
+\lambda_{s}\mathcal{L}_{\mathrm{nmpjpe}}
+\lambda_{v}\mathcal{L}_{\mathrm{vel}}
+\lambda_{d}\mathcal{L}_{\mathrm{diff}}
+\lambda_{lb}\mathcal{L}_{\mathrm{lb}},
\end{equation}
where $\mathcal{L}_{\mathrm{mpjpe}}$ and $\mathcal{L}_{\mathrm{nmpjpe}}$ measure the pose reconstruction error using MPJPE and normalised MPJPE, respectively. $\mathcal{L}_{\mathrm{vel}}$ supervises motion consistency through velocity differences, while $\mathcal{L}_{\mathrm{diff}}$ regularises consecutive-frame variations to encourage temporally smooth predictions~\cite{liu2025tcpformer}. We set $\lambda_{s}=0.5$, $\lambda_{v}=20$, $\lambda_{d}=0.5$ in all experiments. Since our fusion module assigns node-wise weights to different graph pathways, we further apply a lightweight load-balancing regularisation to avoid one pathway dominating the fusion process:
\begin{equation}
\mathcal{L}_{\mathrm{lb}}
=
\sum_{r=1}^{K}
\left(
\frac{1}{|\Omega|}
\sum_{i\in\Omega}
\omega_{r}^{i}
-
\frac{1}{K}
\right)^2,
\end{equation}
where $K$ denotes the number of graph pathways, $\Omega$ is the set of all joint-time nodes in a mini-batch, and $\omega_{r}^{i}$ is the fusion weight assigned to pathway $r$ for node $i$.

\section{Experiments}
\label{sec:experiment}

\subsection{Experimental Setup}

\subsubsection{Datasets} 
We evaluate the proposed HSTGFormer on two standard benchmarks for monocular 3D human pose estimation: Human3.6M~\cite{h36m} and MPI-INF-3DHP~\cite{3dhp}. These two datasets provide complementary evaluation settings, with Human3.6M mainly focusing on controlled indoor scenarios and MPI-INF-3DHP covering more diverse indoor and outdoor environments.

\noindent\textbf{Human3.6M} is a widely used 3D human pose estimation benchmark with 3.6M annotated frames from 11 subjects performing 15 actions in indoor settings. Following~\cite{liu2025tcpformer,li2026MASC,mehraban2024motionagformer}, we train on S1, S5, S6, S7, and S8, and evaluate on S9 and S11. We report Mean Per-Joint Position Error (MPJPE) and Procrustes-aligned MPJPE (P-MPJPE) as evaluation metrics.

\noindent\textbf{MPI-INF-3DHP} is a more challenging benchmark containing diverse indoor and outdoor motions with larger variations in viewpoints, poses, and environmental conditions, making it suitable for evaluating the generalisation ability. Following \cite{liu2025tcpformer,li2026MASC,mehraban2024motionagformer}, we report MPJPE, Percentage of Correct Keypoints (PCK), and Area Under the Curve (AUC).

\subsubsection{Implementation Details}
Our framework is implemented in PyTorch and trained end-to-end using AdamW with a weight decay of $0.01$. The feature dimension is set to $128$.

\noindent\textbf{Human3.6M}: We train for $80$ epochs with batch size $6$ and sequence length $243$. The initial learning rate is $5\times10^{-4}$ with an exponential decay factor of $0.99$. The short/long ADSTG windows are set to $27/81$ frames. Following~\cite{mehraban2024motionagformer,liu2025tcpformer}, we evaluate with both ground-truth 2D keypoints and Stacked Hourglass detections~\cite{newell2016stackedhourglass}.

\noindent\textbf{MPI-INF-3DHP}: We train for $90$ epochs with batch size $6$ and sequence length $81$. The initial learning rate and decay follow Human3.6M. The short/long ADSTG windows are set to $9/27$ frames. Following~\cite{mehraban2024motionagformer,liu2025tcpformer}, we use ground-truth 2D inputs for evaluation.

\subsection{Quantitative Comparisons with State-of-the-Art Methods}

In this section, we quantitatively compare the proposed HSTGFormer with recent state-of-the-art methods on Human3.6M and MPI-INF-3DHP. The comparisons include a broad range of Transformer-based and graph-enhanced pose estimation approaches. Notably, MotionAGFormer~\cite{mehraban2024motionagformer} and TCPFormer~\cite{liu2025tcpformer} use the same backbone encoder as our method, enabling a fair comparison of the proposed graph reasoning modules.

\begin{table}[t]
\setlength{\tabcolsep}{1.4mm} 
\centering
\small
\renewcommand{\arraystretch}{1.1}
\resizebox{\textwidth}{!}{%
\begin{tabular}{l | c | c c | c c | c c c}
\toprule
\textbf{Method} & \textbf{Venue} & \textbf{CE} & $\textbf{T}$ & \textbf{Parameter} & \textbf{MACs} & \textbf{MACs/frames}$\downarrow$ & \textbf{MPJPE}$\downarrow$ & \textbf{P-MPJPE}$\downarrow$ \\
\midrule
MHFormer \cite{li2022mhformer} & CVPR'22 & \cmark & 351 & 30.9M & 7.1G & 7096M & 43.0 & 34.4 \\
MixSTE \cite{zhang2022mixste} & CVPR'22 & \xmark & 243 & 33.6M & 139.0G & 572M  & 40.9 & 32.6 \\
P-STMO \cite{shan2022pstmo}& ECCV'22 & \cmark & 243 & 6.2M  & 0.7G   & 740M  & 42.8 & 34.4  \\
PoseFormerV2 \cite{zhao2023poseformerv2} & CVPR'23 & \cmark & 243 & 14.3M & 0.5G & 528M & 45.2 & 35.6  \\
GLA-GCN \cite{yu2023glagcn}& ICCV'23 & \cmark & 243 & 1.3M & 1.5G & 1556M & 44.4 & 34.8 \\
MotionBERT \cite{zhu2023motionbert} & ICCV'23 & \xmark & 243 & 42.3M & 174.8G & 719M & 39.2 & 32.9 \\
\rowcolor{blue!5} MotionAGFormer-L \cite{mehraban2024motionagformer} & WACV'24 & \xmark & 243 & 19.0M & 78.3G & 322M & 38.4 & 32.5\\
PoseRetNet \cite{zheng2024PoseRetNet}& ECCV'24 & \xmark & 243 & 25.2M & 104.5G & 430M & 40.4 & 32.5\\
KTPFormer \cite{peng2024ktpformer} & CVPR'24 & \xmark & 243 & 33.7M & 69.5G & \underline{286M} & 40.1 & 31.9\\
H2OT$+$MotionAGFormer \cite{li2025h2ot} & TPAMI'25 & \xmark & 243 & 11.7M & 38.9G & - & \underline{38.5} & - \\
\rowcolor{blue!5} TCPFormer \cite{liu2025tcpformer} & AAAI'25 & \xmark & 243 & 35.1M & 109.2G & 449M & \textbf{37.9} & \underline{31.7} \\
\midrule
\rowcolor{blue!5} Ours & -- & \xmark & 243 & 14.2M & 58.5G & \textbf{240M} & \textbf{37.9} & \textbf{31.5} \\
\bottomrule
\end{tabular}
}
\caption{Results on Human3.6M in millimeters under MPJPE and P-MPJPE. CE indicates centre-frame prediction. $T$ is the number of input frames. MACs/frame represents the number of multiply-accumulate operations per output frame, where a lower value indicates higher computational efficiency. \textbf{Bold}/\underline{underline} indicates the best/second-best result. Blue rows indicate methods sharing the same backbone encoder.}
\label{tab:h36m}
\end{table}

\noindent\textbf{Indoor Monocular Human3.6M Dataset.}
As shown in Table~\ref{tab:h36m}, our method achieves highly competitive performance on Human3.6M, obtaining the best P-MPJPE of $31.5$ mm and matching the best MPJPE of $37.9$ mm among all compared methods. Compared with KTPFormer, our model reduces MPJPE from $40.1$ mm to $37.9$ mm and P-MPJPE from $31.9$ mm to $31.5$ mm, corresponding to relative improvements of $5.5\%$ and $1.3\%$, respectively. Compared with MotionAGFormer-L, which shares the same backbone encoder, our method improves MPJPE from $38.4$ mm to $37.9$ mm and P-MPJPE from $32.5$ mm to $31.5$ mm. These results show that HSTGFormer improves pose estimation mainly through more effective graph-enhanced spatial-temporal reasoning, rather than simply increasing model capacity.

\noindent\textbf{Efficiency Comparison.}
Beyond accuracy, our method achieves a favourable accuracy-efficiency trade-off. Compared with TCPFormer, which uses the same backbone encoder, our method obtains comparable MPJPE and better P-MPJPE with $59.5\%$ fewer parameters and $46.5\%$ lower MACs/frame. It also reduces parameters and MACs/frame by $25.3\%$ and $25.5\%$ compared with MotionAGFormer-L.

\noindent\textbf{In-the-wild MPI-INF-3DHP Datasets.} 
Table~\ref{tab:3dhp} reports the quantitative results on MPI-INF-3DHP, which contains more diverse scenes and larger motion variations than Human3.6M. Our method achieves the best overall performance. Compared with TCPFormer, which shares the same backbone encoder, our method improves AUC from $87.7$ to $89.3$ and reduces MPJPE from $15.0$ mm to $14.0$ mm (-6.7\%). Compared with MotionAGFormer-L, our method improves AUC by $4.0$ points and reduces MPJPE by $13.6\%$. These gains on the more challenging MPI-INF-3DHP benchmark indicate that the proposed HSTG and ADSTG modules improve robustness to diverse motion patterns and scene variations.

\begin{table}[t]
\setlength{\tabcolsep}{1.6mm}
\centering
\small
\renewcommand{\arraystretch}{1.1}
\resizebox{0.7\textwidth}{!}{%
\begin{tabular}{l | c | c c | c c c}
\toprule
\textbf{Method} & \textbf{Venue} & \textbf{CE} & $\textbf{T}$ & \textbf{PCK}$\uparrow$ & \textbf{AUC}$\uparrow$ & \textbf{MPJPE}$\downarrow$ \\
\midrule
MHFormer \cite{li2022mhformer} & CVPR'22 & \cmark & 9 & 93.8 & 63.3 & 58.0 \\
MixSTE \cite{zhang2022mixste} & CVPR'22 & \xmark &  27 & 94.4 & 66.5 & 54.9 \\
P-STMO \cite{shan2022pstmo} & ECCV'22 & \cmark & 81 & 97.9 &  75.8 & 32.2 \\
PoseFormerV2 \cite{zhao2023poseformerv2} & CVPR'23 & \cmark &  81 & 97.9 & 78.8 & 27.8 \\
GLA-GCN \cite{yu2023glagcn} & ICCV'23 & \cmark &  81 & 98.5 & 79.1 & 27.7 \\
MotionBERT \cite{zhu2023motionbert} & ICCV'23 & \xmark &  - & - & - & - \\
\rowcolor{blue!5} MotionAGFormer-L \cite{mehraban2024motionagformer} & WACV'24 & \xmark & 81 & 98.2 & 85.3 & 16.2 \\
PoseRetNet \cite{zheng2024PoseRetNet} & ECCV'24 & \xmark & 81 & \textbf{99.1} & 84.4 &  22.2 \\
KTPFormer \cite{peng2024ktpformer} & CVPR'24 & \xmark & 81 & 98.9 & 85.9 & 16.7 \\
H2OT+MotionAGFormer \cite{li2025h2ot} & TPAMI'25 & \xmark & 81 & \textbf{99.1} & 85.2 & 18.0 \\
\rowcolor{blue!5} TCPFormer \cite{liu2025tcpformer} & AAAI'25 & \xmark & 81 & \underline{99.0} & \underline{87.7} & \underline{15.0} \\
\midrule
\rowcolor{blue!5} Ours & -- & \xmark & 81 & \textbf{99.1} & \textbf{89.3} & \textbf{14.0} \\
\bottomrule
\end{tabular}
}
\setlength{\abovecaptionskip}{0.1cm}
\caption{Results on MPI-INF-3DHP. CE indicates centre-frame prediction. $T$ denotes the number of input frames. \textbf{Bold}/\underline{underline} indicates the best/second-best result. Blue rows indicate methods sharing the same backbone encoder.}
\label{tab:3dhp}
\end{table}

\subsection{Ablation Studies and Analysis}

\subsubsection{Main Component Ablation Studies}

\begin{table}[t]
\centering
\setlength{\tabcolsep}{5pt}
\small
\renewcommand{\arraystretch}{1.12}
\resizebox{0.8\linewidth}{!}{%
\begin{tabular}{lcc}
\toprule
\textbf{Variant} & \textbf{MPJPE}$\downarrow$ & \textbf{P-MPJPE}$\downarrow$ \\
\midrule
(1) Baseline (w/o HSTG \& ADSTG) & 41.1 & 36.0 \\
(2) w/o HSTG                 & 39.2 & 32.9 \\
(3) w/o ADSTG                & 38.8 & \underline{32.5} \\
(4) w/o Node-wise Adaptive Fusion (fixed $0.5/0.5$ weights)            & \underline{38.3} & 32.6 \\
(5) w/o load-balance loss  ($\mathcal{L}_{\mathrm{lb}}$)    & 38.8 & 33.0 \\
\midrule
\textbf{HSTGFormer (Ours)} & \textbf{37.9} & \textbf{31.5} \\
\bottomrule
\end{tabular}
}
\setlength{\abovecaptionskip}{0.1cm}
\caption{Main component ablation on Human3.6M.
We evaluate the contribution of the proposed HSTG, ADSTG, adaptive fusion strategy, and load-balance loss.
Results are reported in MPJPE and P-MPJPE (mm).
\textbf{Bold}/\underline{underline} indicates the best/second-best result.}
\label{tab:main_ablation}
\end{table}

Table~\ref{tab:main_ablation} presents the main ablation study on Human3.6M. We evaluate the contribution of the two proposed graph reasoning modules, HSTG and ADSTG, as well as the adaptive node-wise fusion strategy. Variants (1)-(3) mainly examine the effectiveness of the graph reasoning design. Compared with the baseline without HSTG and ADSTG, adding either component brings clear improvements, reducing MPJPE from $41.1$ mm to $39.2$ mm and $38.8$ mm, respectively. This shows that both local coupled spatial-temporal reasoning and adaptive temporal graph modelling contribute to better pose estimation. Variants (4)-(5) further analyse the fusion mechanism. Replacing node-wise adaptive fusion with fixed $0.5/0.5$ weights degrades MPJPE from $37.9$ mm to $38.3$ mm, indicating that different joint-time nodes benefit from different graph contexts. Removing the load-balance loss also weakens the performance, suggesting that balanced utilization of the two branches helps stabilize adaptive fusion. Overall, the full HSTGFormer achieves the best MPJPE and P-MPJPE, showing that HSTG and ADSTG provide complementary spatial-temporal representations, while the proposed node-wise fusion further improves adaptive spatial-temporal reasoning.

\begin{table}[t]
\centering
\setlength{\tabcolsep}{2.3pt}
\small
\renewcommand{\arraystretch}{1.0}
\resizebox{0.7\linewidth}{!}{%
\begin{tabular}{l | l | c c}
\toprule
\textbf{Component} & \textbf{Variant} & \textbf{MPJPE}$\downarrow$ & \textbf{P-MPJPE}$\downarrow$ \\
\midrule
\multirow{3}{*}{HSTG} 
& (1) w/ Factorised MLP & 39.7 & 33.0 \\
& (2) w/ Factorised GCN  & \underline{38.1} & \underline{32.1} \\
& \textbf{Ours} & \textbf{37.9} & \textbf{31.5} \\
\midrule
\multirow{3}{*}{ADSTG}
& (1) w/ Fixed Dual-Scale Temporal Convolution & 39.1 & 32.7 \\
& (2) w/ Fixed Temporal Attention & \underline{38.7} & \underline{32.6} \\
& \textbf{Ours} & \textbf{37.9} & \textbf{31.5} \\
\bottomrule
\end{tabular}%
}
\setlength{\abovecaptionskip}{0.1cm}
\caption{Ablation studies on HSTG and ADSTG design choices. Results are reported in MPJPE and P-MPJPE (mm). \textbf{Bold}/\underline{underline} indicate the best/second-best results.}
\label{tab:abl_graph_design}
\end{table}

\begin{table}[t]  
  \centering
  \setlength{\tabcolsep}{2.2pt}
  \footnotesize
  \renewcommand{\arraystretch}{1.5}
  \resizebox{\columnwidth}{!}{%
  \begin{tabular}{l|c|c|ccccccccccccccc}
    \toprule
    \textbf{Method} & \textbf{Venue} & \textbf{T} & Dir. & Disc. & Eat & Greet & Phone & Photo & Pose & Pur. & Sit & SitD. & Smoke & Wait & WalkD. & Walk & WalkT. \\
    \midrule
MHFormer~\cite{li2022mhformer} & CVPR'22 & 351 & 31.5 & 34.9 & 32.8 & 33.6 & 35.3 & 39.6 & 32.0 & 32.2 & 43.5 & 48.7 & 36.4 & 32.6 & 34.3 & 23.9 & 25.1 \\
MixSTE~\cite{zhang2022mixste} & CVPR'22 & 243 & 32.0 & 34.2 & 31.7 & 33.7 & 34.4 & 39.2 & 32.0 & 31.8 & 42.9 & 46.9 & 35.5 & 32.0 & 34.4 & 23.6 & 25.2 \\
P-STMO~\cite{shan2022pstmo} & ECCV'22 & 243 & 31.3 & 35.2 & 32.9 & 33.9 & 35.4 & 39.3 & 32.5 & 31.5 & 44.6 & 48.2 & 36.3 & 32.9 & 34.4 & 23.8 & 23.9 \\
GLA-GCN~\cite{yu2023glagcn} & ICCV'23 & 243 & 32.4 & 35.3 & 32.6 & 34.2 & 35.0 & 42.1 & 32.1 & 31.9 & 45.5 & 49.5 & 36.1 & 32.4 & 35.6 & 23.5 & 24.7 \\
\rowcolor{blue!5} MotionAGFormer-L~\cite{mehraban2024motionagformer} & WACV'24 & 243 & 31.0 & 32.6 & \underline{31.1} & 28.0 & 34.0 & 38.8 & 31.5 & 30.1 & 41.4 & \underline{45.5} & 34.9 & 30.8 & 31.3 & 22.8 & 23.2 \\
KTPFormer~\cite{peng2024ktpformer} & CVPR'24 & 243 & \underline{30.1} & 32.3 & \textbf{29.6} & 30.8 & \textbf{32.3} & \textbf{37.3} & 30.0 & \underline{30.2} & \underline{41.0} & \textbf{45.3} & \textbf{33.6} & \underline{29.9} & 31.4 & \underline{21.5} & 22.6 \\
PoseRetNet~\cite{zheng2024PoseRetNet} & ECCV'24 & 243 & 30.8 & 33.1 & 31.3 & 31.8 & 33.4 & 37.7 & 30.1 & 30.5 & 43.4 & \underline{45.5} & \underline{34.3} & 30.3 & 31.5 & \textbf{21.4} & 22.7 \\
\rowcolor{blue!5} TCPFormer~\cite{liu2025tcpformer} & AAAI'25 & 243 & \underline{30.1} & \textbf{31.6} & 31.4 & \textbf{27.3} & 33.5 & \underline{37.6} & \textbf{29.4} & 29.6 & 41.1 & 45.9 & 34.4 & \textbf{29.6} & \underline{30.6} & 21.7 & \underline{22.3} \\
\midrule
\rowcolor{blue!5}
\textbf{Ours} & - & 243 & \textbf{29.9} & \underline{31.9} & \textbf{29.6} & \underline{27.7} & \underline{33.3} & \underline{37.6} & \underline{29.5} & \textbf{29.3} & \textbf{39.6} & \textbf{45.3} & 34.8 & 30.0 & \textbf{30.2} & 22.3 & \textbf{22.1} \\
\bottomrule
  \end{tabular}%
  }
  \caption{Results on Human3.6M under P-MPJPE for 15 actions. \textbf{Bold}/\underline{underline} indicates the best/second-best result. Blue rows indicate methods sharing the same backbone encoder.}
  \label{tab:h36m_p2}
\end{table}

\subsubsection{Graph Design Analysis} 

Table~\ref{tab:abl_graph_design} explores the design choices of HSTG and ADSTG by replacing them with alternative architectures. For HSTG, factorised MLPs lead to a clear performance drop, with our design reducing MPJPE by $4.5\%$ and P-MPJPE by $4.2\%$, indicating that point-wise transformations are insufficient for modelling structured joint dependencies and coupled spatial-temporal interactions. Factorised GCNs perform better than MLPs but still underperform our design, suggesting that adaptive graph attention enables more flexible spatial-temporal aggregation. For ADSTG, replacing adaptive temporal graph modelling with fixed dual-scale temporal convolutions or fixed temporal attention also degrades performance. This shows that content-adaptive temporal neighbourhoods are more effective than predefined temporal kernels.

\subsubsection{Per-action Quantitative Analysis}

To better understand the action-wise behaviour of HSTGFormer, we analyse the per-action performance on Human3.6M in Table~\ref{tab:h36m_p2}. Our method achieves the best results on several representative actions, including \textit{Direction}, \textit{Eating}, \textit{Purchases}, \textit{Sitting}, \textit{Sitting Down}, \textit{Walking Dog}, and \textit{Walking Together}, while remaining competitive on \textit{Discussion}, \textit{Phone}, \textit{Photo}, and \textit{Pose}. The gains are particularly clear on complex or motion-intensive actions, suggesting that HSTGFormer effectively captures continuous spatial-temporal dependencies.

\subsection{Qualitative Analysis}
\subsubsection{Pose Estimation Visualisation}
\begin{figure}[t]
\centering
  \includegraphics[width=0.7\linewidth]{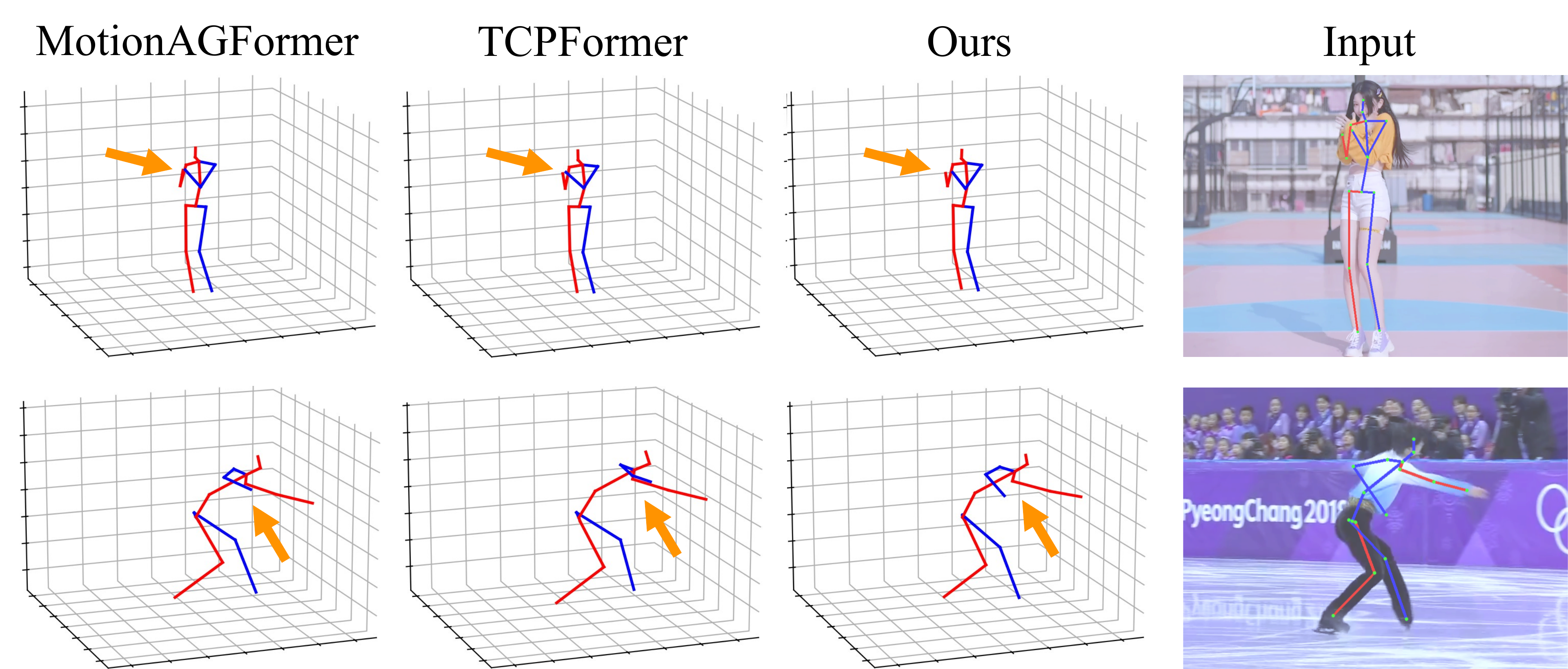}
  \caption{Qualitative comparison against TCPFormer~\cite{liu2025tcpformer} and MotionAGFormer \cite{mehraban2024motionagformer} on in-the-wild videos. Orange arrows highlight challenging pose regions.}
  \label{fig:quali_wild}
\end{figure}

\begin{figure}[t]
\centering
  \includegraphics[width=0.75\linewidth]{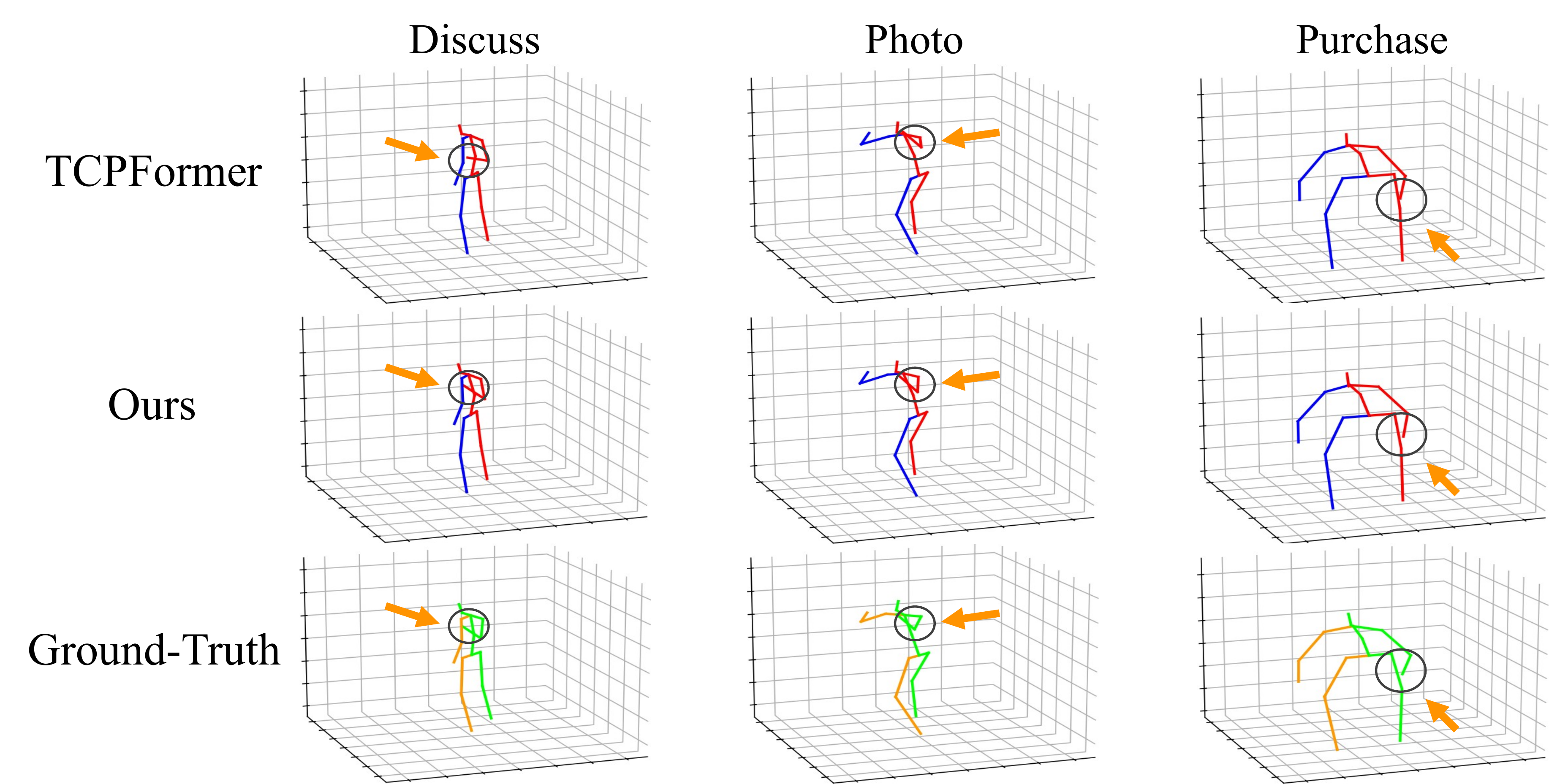}
  \caption{Qualitative comparison against TCPFormer~\cite{liu2025tcpformer} on the Human3.6M dataset. Orange arrows highlight challenging pose regions.}
  \label{fig:quali_h36m}
\end{figure}

Figures~\ref{fig:quali_wild} and~\ref{fig:quali_h36m} present qualitative comparisons on both in-the-wild videos and the Human3.6M dataset. Compared with MotionAGFormer and TCPFormer, our method produces more structurally consistent and anatomically plausible 3D poses under challenging body articulations and self-occlusions. In particular, our predictions better preserve local joint geometry and long-range body coordination in difficult regions such as arms, shoulders, and torso bending poses. These qualitative results further demonstrate the effectiveness of the proposed HSTG and ADSTG modules for robust spatial-temporal pose reasoning.

\begin{figure}[t]
\centering
  \includegraphics[width=0.89\linewidth]{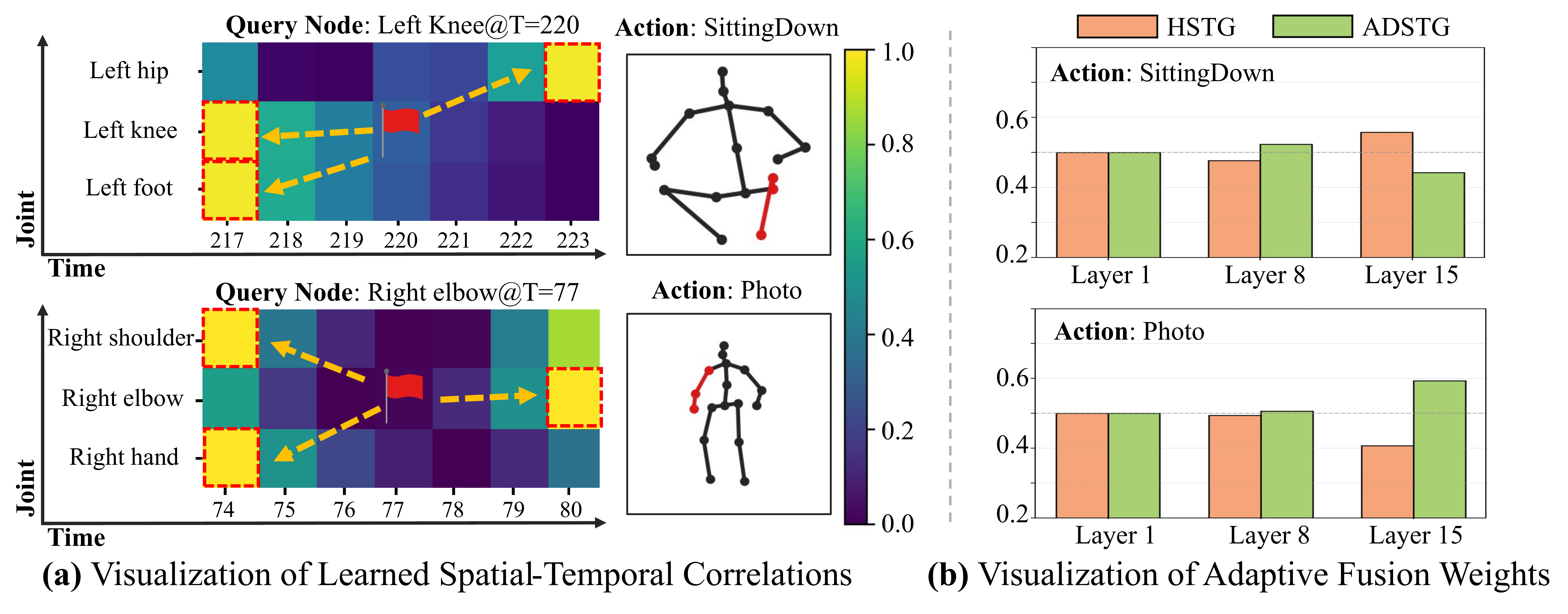}

  \caption{\textbf{(a)} Visualisation of learned spatial-temporal correlations in HSTG. 
Highlighted regions indicate that the query node attends to anatomically related joints across nearby timesteps.
\textbf{(b)} Visualisation of adaptive fusion weights across different layers and actions.
}
  \label{fig:quali_ablation}
\end{figure}

\begin{figure}[t]
\centering
  \includegraphics[width=\linewidth]{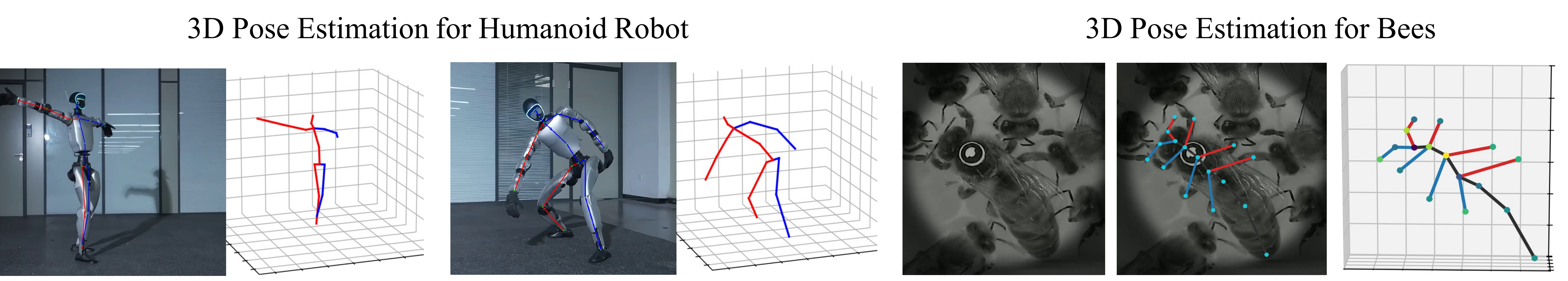}

\caption{Zero-shot qualitative visualisations on humanoid robots and bees using the Human3.6M-trained checkpoint.}
\label{fig:quali_robotbee}
\end{figure}

\subsubsection{Correlation Weights Analysis for HSTG}
Figure~\ref{fig:quali_ablation} (a) visualizes the learned spatial-temporal correlation weights in HSTG. For each query joint-time node, we show its correlation responses to anatomically related joints within a local temporal neighbourhood. The highlighted regions indicate that the most relevant information does not always come from the same frame. Instead, the query node can attend to related joints across nearby timesteps, such as the knee receiving strong responses from neighbouring hip/foot joints at adjacent frames. This observation supports our motivation that local pose dynamics are inherently coupled in both spatial and temporal dimensions. By extending the skeleton graph into localised temporal neighbourhoods, HSTG can capture such cross-frame structural dependencies while still preserving anatomical constraints.

\subsubsection{Adaptive Fusion Weights Analysis}
Figure~\ref{fig:quali_ablation} (b) shows the adaptive fusion weights assigned to HSTG and ADSTG across different layers and actions. The weights are not fixed, but vary with both network depth and motion context. For \textit{SittingDown}, the model assigns relatively higher HSTG weights in deeper layers, suggesting that local coupled spatial-temporal reasoning becomes more important for complex body articulation. In contrast, for \textit{Photo}, the ADSTG weight increases in deeper layers, indicating a stronger reliance on adaptive temporal dependency modelling.

\subsubsection{Broader Impact Discussion}

Figure~\ref{fig:quali_robotbee} shows qualitative zero-shot visualisations beyond human pose estimation using the Human3.6M-trained checkpoint without additional fine-tuning. Although the proposed framework is trained only on human pose data, it can produce visually coherent 3D structures for articulated humanoid robots and bees. These preliminary examples are not intended as quantitative validation, but rather illustrate the potential of graph-enhanced spatial-temporal reasoning for broader articulated structure modelling.

\section{Conclusion}
\label{sec:conclusion}

In this paper, we propose HSTGFormer, a graph-enhanced Transformer framework for efficient monocular 3D human pose estimation. Unlike conventional spatial-then-temporal modelling strategies, the proposed framework reformulates spatial-temporal reasoning from a graph perspective through two complementary modules: the Hyper Spatial-Temporal Graph (HSTG) and the Adaptive Dual-Scale Temporal Graph (ADSTG). HSTG extends conventional skeleton graphs into localised temporal neighbourhoods to enable efficient coupled spatial-temporal reasoning, while ADSTG performs adaptive temporal dependency modelling over complementary temporal ranges. In addition, a lightweight node-wise fusion mechanism dynamically integrates the two graph representations according to different motion contexts. Extensive experiments on Human3.6M and MPI-INF-3DHP demonstrate that HSTGFormer achieves strong accuracy while maintaining high computational efficiency and low memory cost. Further ablation studies and qualitative analyses verify the effectiveness of the proposed graph reasoning design. Preliminary zero-shot results on humanoid robots and bees also suggest its potential applicability beyond human pose estimation.

\section*{Acknowledgements}
This work was supported by EU H2020-FET RoboRoyale project (964492).

\bibliography{egbib}
\end{document}

\maketitle

In this supplementary material, we provide additional experimental results and analyses to further validate the effectiveness of the proposed HSTGFormer framework.

%-------------------------------------------------------------------------
\section{Additional Experiment Results}
\subsection{Effect of HSTG Temporal Neighbourhood Radius}

\begin{table}[ht]
\centering
\setlength{\tabcolsep}{10pt}
\begin{tabular}{c|c|c c}
\hline
$w$ & Temporal Range & MPJPE$\downarrow$ (mm) & P-MPJPE$\downarrow$ (mm)\\
\hline
1 & $t\!\pm\!1$ \; (3 frames) & 38.4                & 32.2 \\
\rowcolor{blue!5} 3 & $t\!\pm\!3$ \; (7 frames) & \textbf{37.9}       & \textbf{31.5} \\
5 & $t\!\pm\!5$ \; (11 frames) & \underline{38.2}               & \underline{31.9} \\
7 & $t\!\pm\!7$ \; (15 frames) & 38.7               & 32.4 \\
9 & $t\!\pm\!9$ \; (19 frames) & 38.9               & 32.8 \\
\hline
\end{tabular}
\setlength{\abovecaptionskip}{0.1cm}
\caption{Effect of HSTG temporal neighbourhood radius $w$ on Human3.6M. 
$w$ denotes the number of neighbouring timesteps on each side of a joint-time node. 
\textbf{Bold}/\underline{underline} indicate the best/second-best results. The blue row represents our configuration.}
\label{tab:window}
\end{table}

\noindent Table~\ref{tab:window} analyses the effect of the temporal neighbourhood radius $w$ in HSTG. Smaller neighbourhoods (e.g., $w=1$) provide insufficient temporal context for local spatial-temporal reasoning, leading to inferior performance. Increasing the neighbourhood radius improves performance by enabling each joint-time node to aggregate richer local motion dependencies across nearby frames. The best performance is achieved at $w=3$, corresponding to a local temporal range of seven frames centred at each joint-time node. However, further enlarging the temporal neighbourhood gradually degrades performance, suggesting that excessively large receptive fields may introduce redundant temporal context and weaken localized motion modelling. These results demonstrate the importance of maintaining an appropriate local spatial-temporal receptive field in HSTG.

\subsection{Effect of ADSTG Temporal Window}

\begin{table}[ht]
\centering
\setlength{\tabcolsep}{10pt}
\begin{tabular}{c|c|c c}
\hline
Variants & Temporal Window & MPJPE$\downarrow$ (mm) & P-MPJPE$\downarrow$ (mm)\\
\hline
Short-range only & $27/27$ & \underline{38.6} & \underline{32.4} \\
Long-range only & $81/81$ & 38.8 & 32.9 \\
\rowcolor{blue!5} Ours (ADSTG) & $27/81$ & \textbf{37.9} & \textbf{31.5} \\
\hline
\end{tabular}
\setlength{\abovecaptionskip}{0.1cm}
\caption{Effect of adaptive dual-scale temporal modelling in ADSTG on Human3.6M with sequence length $T=243$. 
The short- and long-range temporal windows are set to $27$ and $81$ frames, respectively. \textbf{Bold}/\underline{underline} indicate the best/second-best results. The blue row represents our configuration.}
\label{tab:adstg}
\end{table}

\noindent Table~\ref{tab:adstg} shows the effect of adaptive dual-scale temporal modelling in ADSTG. Using only short-range temporal windows captures fine-grained local motion dynamics but lacks sufficient long-range context, while relying solely on long-range windows reduces sensitivity to local motion variations. 
By adaptively combining complementary short- and long-range temporal graphs, ADSTG achieves the best performance in both MPJPE and P-MPJPE. 

\section{More Qualitative Results in In-The-Wild Scenarios}

\begin{figure}[t]
\centering
  \includegraphics[width=\linewidth]{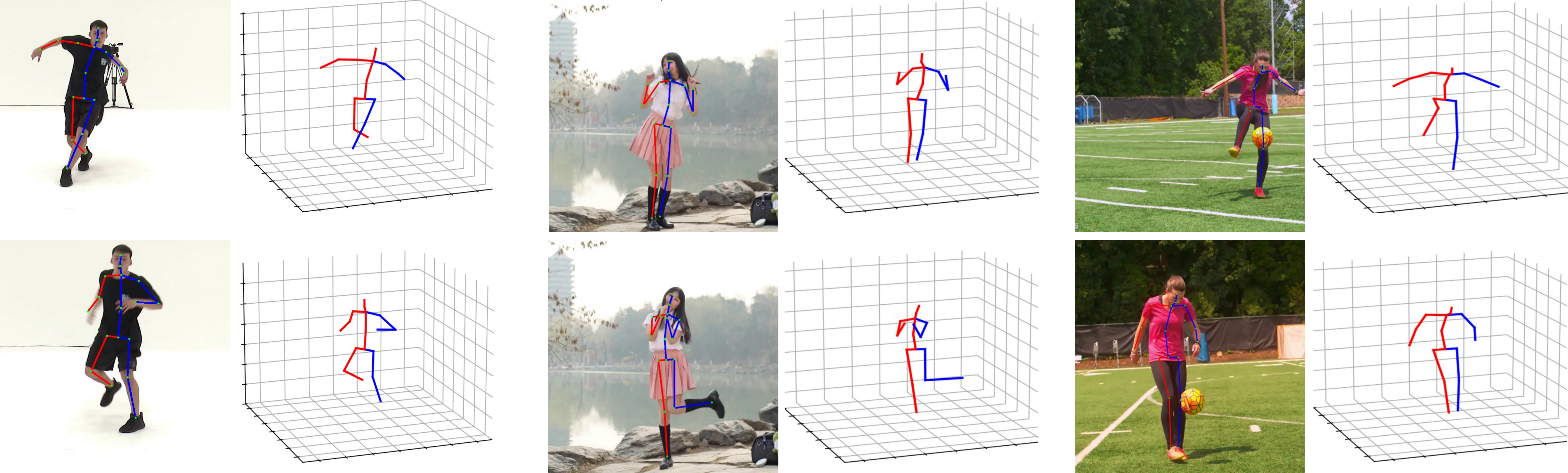}
\caption{Qualitative visualisation results on in-the-wild videos. 
For each example, we show the input image with 2D pose and the corresponding predicted 3D pose produced by HSTGFormer.}
  \label{fig:quali_wild}
\end{figure}

\noindent To further evaluate the qualitative behaviour of HSTGFormer in unconstrained scenarios, we visualise predicted 3D poses on in-the-wild videos \cite{aist2019music} in Figure~\ref{fig:quali_wild}. Although the scenes contain diverse appearances, viewpoints, and motion patterns, HSTGFormer produces structurally plausible 3D poses that preserve the overall body configuration. These results suggest that the proposed graph-enhanced spatial-temporal reasoning can generalise to challenging in-the-wild settings beyond controlled training scenarios.

\bibliography{egbib}